%% file: main.tex
\documentclass[sigconf]{acmart}
\makeatletter
\@ifpackageloaded{hyperxmp}{\relax}{\RequirePackage{hyperref}}
\makeatother

\usepackage{multirow}

\AtBeginDocument{%
  \providecommand\BibTeX{{%
    \normalfont B\kern-0.5em{\scshape i\kern-0.25em b}\kern-0.8em\TeX}}}

\acmBooktitle{Proceedings of the 2026 ACM/IEEE International Conference on Human-Robot Interaction (HRI '26), March 16--19, 2026, Edinburgh, Scotland, UK}

\begin{document}

\title{Cognitive Trust in HRI: ``Pay Attention to Me and I'll Trust You Even if You are Wrong''}



    \author{Adi Manor}
\email{adi.manor@milab.runi.ac.il}
\orcid{0009-0005-2117-5867} 
\affiliation{Faculty of Data and Decision Sciences, 
Technion;
  \institution{milab, Reichman University}
  \streetaddress{8 Ha'universita Street}
  \city{Haifa; Herzliya}
  \country{Israel} 
}

\author{Dan Cohen}
\email{Dan.Cohen@milab.runi.ac.il}
\orcid{0009-0003-3385-4221} 
\affiliation{milab,
  \institution{Reichman University}
  \streetaddress{8 Ha'universita Street}
  \city{Herzliya}
  \country{Israel} 
}
\author{Ziv Keidar}
\email{Ziv.Keidar@milab.runi.ac.il}
\orcid{0009-0000-9490-5158} 
\affiliation{milab,
  \institution{Reichman University}
  \streetaddress{8 Ha'universita Street}
  \city{Herzliya}
  \country{Israel} 
}

    \author{Avi Parush}
\email{aparush@technion.ac.il}
\orcid{0000-0003-4435-8576} 
\affiliation{Faculty of Data and Decision Sciences,
  \institution{Technion}
  \streetaddress{8 Ha'universita Street}
  \city{Haifa}
  \country{Israel} 
}
    
\author{Hadas Erel}
\email{hadas.erel@milab.runi.ac.il}
\orcid{0000-0001-7099-6104} 
\affiliation{milab,
  \institution{Reichman University}
  \streetaddress{8 Ha'universita Street}
  \city{Herzliya}
  \country{Israel} 
}

  



\renewcommand{\shortauthors}{Manor et al.}


\begin{abstract}

\input{section/0_abstract.tex}

\end{abstract}





\maketitle

\input{section/1_introduction.tex}

\input{section/2_related_work.tex}

\input{section/3_methods.tex}
\input{section/4_Analysis_and_Findings}
\input{section/5_discussion}

\input{section/6_conclusion}

\bibliographystyle{ACM-Reference-Format}
\bibliography{main}

\end{document}

%% file: section/0_abstract.tex
Cognitive trust and the belief that a robot is capable of accurately performing tasks, are recognized as central factors in fostering high-quality human-robot interactions. It is well established that performance factors such as the robot's competence and its reliability shape cognitive trust. Recent studies suggest that affective factors, such as robotic attentiveness, also play a role in building cognitive trust. This work explores the interplay between these two factors that shape cognitive trust. Specifically, we evaluated whether different combinations of robotic competence and attentiveness introduce a compensatory mechanism, where one factor compensates for the lack of the other. In the experiment, participants performed a search task with a robotic dog in a $2\times2$ experimental design that included two factors: competence (high or low) and attentiveness (high or low). The results revealed that high attentiveness can compensate for low competence. Participants who collaborated with a highly attentive robot that performed poorly reported trust levels comparable to those working with a highly competent robot. When the robot did not demonstrate attentiveness, low competence resulted in a substantial decrease in cognitive trust. The findings indicate that building cognitive trust in human-robot interaction may be more complex than previously believed, involving emotional processes that are typically overlooked. We highlight an affective compensatory mechanism that adds a layer to consider alongside traditional competence-based models of cognitive trust. 

%% file: section/1_introduction.tex
\section{Introduction}
As robots rapidly become more complex and capable, the nature of our interactions with them is changing from simple assistance in task execution to complex social and collaborative teamwork. Under these contexts, trust plays a major role in shaping the quality and effectiveness of Human-Robot Interactions (HRI). Trust not only facilitates smoother cooperation and reduces cognitive load, but also influences emotional engagement, willingness to rely on robotic partners, and facilitates long-term acceptance \cite{kok2020trust, hancock2011meta}. When trust is compromised, interactions often become strained, leading to reduced reliance, disengagement, and overall rejection of the robot \cite{naneva2020systematic, lewis2018role}.

As robots become more advanced, our interactions with them are shifting from simple assistance in task performance to a more complex collaborative context, where trust plays a key role in the quality and success of the human-robot interaction (HRI). Trust facilitates smoother cooperation, reduces mental effort, and enhances performance \cite{kok2020trust, cocato2025factors}. It also contributes to emotional engagement, willingness to rely on the robot, and long-term acceptance \cite{hancock2011meta, cocato2025factors}. When trust is broken, interactions often suffer, leading to less reliance, disengagement, and rejection of the robot altogether \cite{naneva2020systematic, lewis2018role}.

HRI studies already indicated that, similar to human relationships, trust in robots has multiple dimensions, each influencing the interaction in distinct ways. Two established dimensions are \textit{Cognitive Trust} (CT) and \textit{Affective trust} (AT) \cite{anzabi2023effect, anzabi2023influence, manor2024attentiveness, manor2025trust, johnson2005cognitive, mcallister1995affect}. CT refers to the belief that the robot is reliable and capable of accurately performing tasks \cite{kok2020trust, salem2015would, hancock2011meta, manor2025trust}. AT, on the other hand, reflects emotional closeness, empathy, and the perception that the robot is supportive and caring \cite{anzabi2023effect, anzabi2023influence, manor2024attentiveness}. Previous studies have identified factors that shape the different trust dimensions \cite{anzabi2023effect, anzabi2023influence, hancock2011meta, cocato2025factors}. Competence was highlighted as the primary factor for establishing CT, with people adjusting their trust and reliance on a robot’s cognitive recommendations according to its reliability and performance \cite{xu2018impact, salem2015would, hancock2011meta}. Conversely, robotic attentiveness, expressed through behaviors such as following, maintaining proximity, and responsiveness, was perceived as a signal of warmth, care, and willingness to help, all of which foster AT \cite{manor2024attentiveness, sidner2004study, moon2014meet}.

\begin{figure} 
    \centering
    \includegraphics[width=0.7\linewidth]{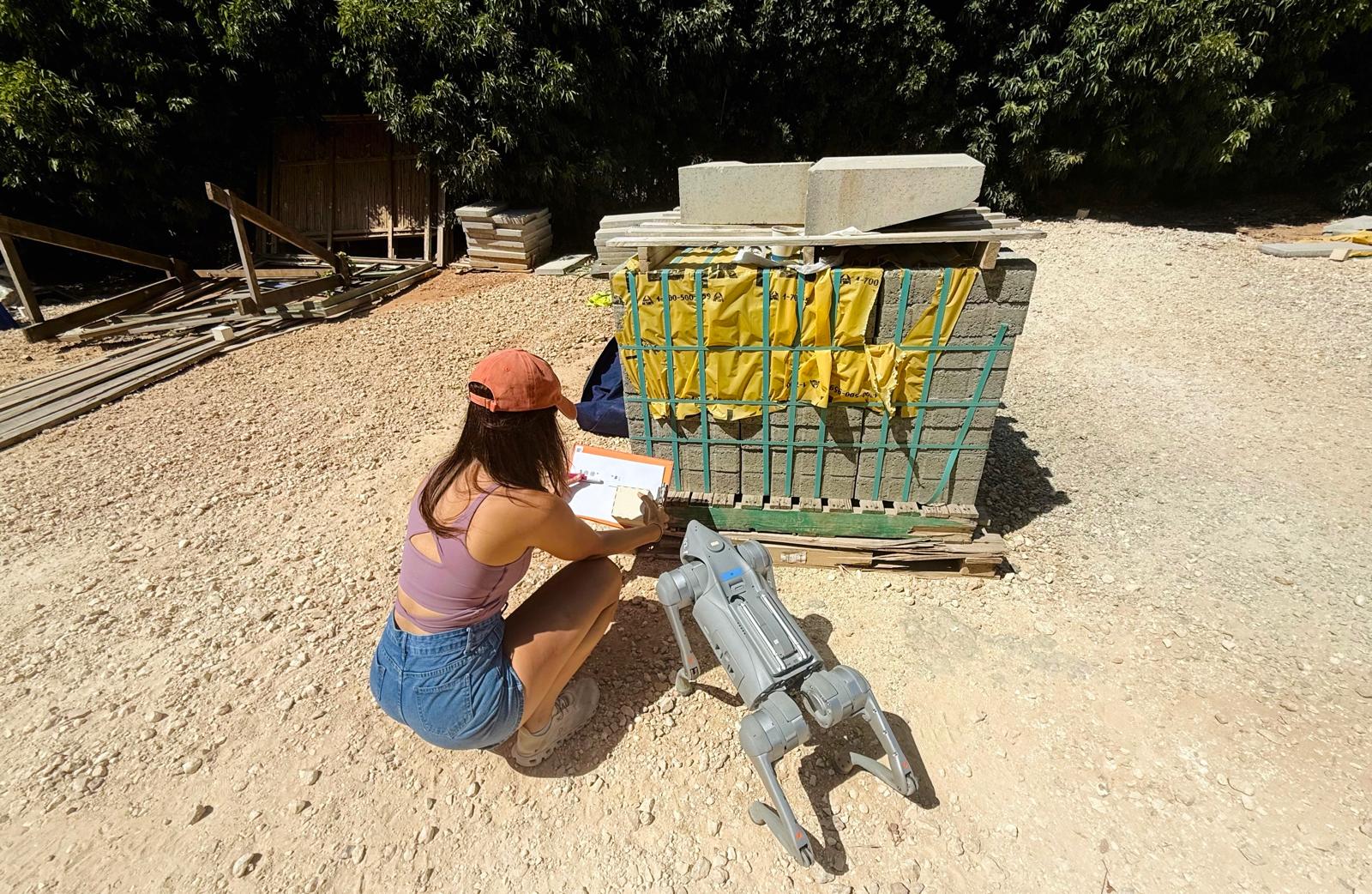}
    \caption{An attentive robotic dog showing interest in a participant performing a search task.}
    \label{fig:DAN}
\end{figure}

Recent studies have shown some overlap between the two trust dimensions and indicate that CT can be shaped either by a robot’s competence or by its perceived attentiveness, suggesting that factors traditionally associated with AT may also contribute to the development of CT \cite{anzabi2023effect, manor2024attentiveness}. This cross-dimensional effect can be explained by people's expectations when interacting with robots \cite{erel2024rosi}, particularly the expectation that a highly competent robot warrants the resources and effort spent on creating a machine capable of assisting humans with certain tasks \cite{robinette2016overtrust}. This common belief that robots are practical machines designed to perform specific tasks does not involve an expectation of emotional or attentive behavior \cite{erel2024rosi}. Thus, when a robot demonstrates attentiveness, it exceeds expectations and is perceived as having additional capabilities beyond basic functionality \cite{erel2024rosi, erel2022carryover, manor2024attentiveness}. This unexpected social behavior can signal greater ability in general, which contributes to CT \cite{manor2024attentiveness}. In contrast, when robots demonstrate high competence alone, they meet expected standards without signaling social-emotional capabilities, thereby limiting their impact to CT only \cite{manor2025trust, robinette2016overtrust}. 

While the surprising impact of robotic attentiveness on CT opens intriguing avenues for shaping people’s trust in robots, it has often been evaluated in isolation, without considering its relationship with competence, which is the primary contributor to CT (i.e., participants had no information regarding the robot's competence \cite{manor2024attentiveness}). This single-factor approach is commonly used when identifying novel phenomena and is typical in the HRI domain which is a rather novel research field in general. However, the case of trust and the impact of both cognitive and affective factors highlight the importance of exploring joint effects and whether they differ from the simple sum of individual effects. Apart from conducting more ecological studies that better represent real-world interactions (that are typically influenced by more than a single factor), identifying the interplay between independent factors and whether they involve compensatory processes may strongly contribute to the flexibility and scope of designing interactions with robots. In the specific case of CT and the two factors known to impact it (i.e., competence and attentiveness), little is known about the relationship between these factors when combined, despite the likely real-world scenario where robots may simultaneously demonstrate different levels of competence and attentiveness. This scenario raises important questions about the mutual effect of competence and attentiveness on CT and whether robot attentiveness can compensate for low competence, resulting in high CT despite the occurrence of robot errors. Identifying such dependencies between different levels of competence and attentiveness can help optimize robot design because their combined effect on CT may differ from their individual effects when tested separately.

To evaluate how the combination of competence and attentiveness affects CT, we designed a collaborative interaction with a robot in which participants performed a task together with a robotic dog. The collaborative context allowed us to test the impact of competence and attentiveness in a setting where CT is highly important for the interaction. We introduced a turn-based search task in which we manipulated the robot's competence and attentiveness in a $2\times2$ experimental design. The robot demonstrated high competence by accurately locating targets related to the search task. It demonstrated low competence by making errors when it was its turn to search. The design of high attentiveness included robotic behaviors such as following the participants during their search, sharing emotional responses, and remaining generally responsive. In contrast, low attentiveness involved not following the participant and being unresponsive to their actions. We measured participants’ level of CT after experiencing interactions that involved different combinations of robotic competence and attentiveness.

%% file: section/2_related_work.tex
\section{Related Work} 
Relevant previous studies include HRI studies exploring cognitive trust, robotic attentiveness, and robot competence.

\subsection{Cognitive trust in HRI} 
Multiple studies have identified factors that shape CT in HRI \cite{gompei2018factors, hancock2011meta}. These studies highlighted robots performance and reliability as primary factors contributing to the development of trust \cite{xu2018impact, nesset2021transparency, salomons2018humans, hancock2011meta}. In contrast, errors (e.g., memory lapses) and physical mistakes (e.g., inaccurate navigation) significantly reduce CT \cite{salem2015would}. More specifically, \citet{xu2018impact} highlighted the importance of robot performance at the beginning of the interaction. Robots that performed well at the beginning of an interaction tended to maintain high CT throughout subsequent interactions, whereas initial errors typically led to a persistent decline in CT \cite{xu2018impact}. Similarly, \citet{desai2012effects} explored how variations in reliability during a navigation task affected CT. They found that participants expressed greater trust when interacting with consistently competent robots, regardless of their performance quality. When a robot's reliability dropped after it demonstrated high competence, trust decreased more drastically in comparison with consistent failures, suggesting that violating expectations had a stronger negative effect than initial low expectations \cite{desai2012effects}. Additionally, several studies suggested that affective robotic behavior can impact CT. \citet{anzabi2023effect} showed that a robot's listening behavior can influence the formation of trust. When the robot's behavior indicated interest and empathy, both AT and CT increased \cite{anzabi2023effect}. \citet{christoforakos2021can} also found that the perception of a robot's warmth can enhance CT. They manipulated robot warmth by designing empathetic, attentive behavior in which the robot celebrated participants’ successes and offered encouragement while they were playing a game. They showed that these empathetic social cues facilitated the development of CT, underscoring the importance of integrating attentiveness cues when building trust in HRI. Similarly, \citet{manor2024attentiveness} found that a simple non-humanoid robot resembling a desk lamp enhanced CT by performing attentive nonverbal gestures (e.g., gazing and leaning toward the participant) when participants shared their plans for the future. Participants explained that robot's attentive behavior implied competence, making the robot seem capable of handling complex tasks.

We extend this line of work by systematically evaluating how the combination of robot competence and attentiveness affects CT. Whereas earlier studies have shown that each factor affects CT on its own, the combined effect of these factors and whether it involves compensatory mechanisms remain largely unexplored.

\subsection{The power of robotic attentiveness}
Previous research highlights the significant impact of robotic attentiveness when integrated into HRI. For example, \citet{johanson2019effect} showed that robotic attentiveness can facilitate engagement. In their study, participants who interacted with a robot that leaned toward them spent more time looking at it, were more attentive and interested, and were more likely to lean forward themselves. The power of attentiveness was also demonstrated in interactions with simple, non-humanoid robots \cite{hoffman2015design, erel2021enhancing, erel2021excluded, manor2024attentiveness}. \citet{manor2022non} designed an interaction with a simple robotic object that demonstrated listening and attentiveness via minimal non verbal gestures that included leaning forward, gazing, and nodding. Participants reported that the interaction with the attentive robot increased their confidence and general sense of security. Using the same robot, \citet{erel2022enhancing} demonstrated that the robot’s attentive gestures can enhance the quality of emotional support provided by participants to one another. The robot that modeled attentiveness using gaze and changes in its body orientation, primed a sense of attentiveness in the interaction, resulting in a sense of warmth and care between the human participants. More generally, Szafir and Mutlu \cite{szafir2012pay} formalized these effects using nonverbal immediacy theory, showing that attentive behaviors such as close proximity, leaning forward, gazing, and body orientation, enhance perceived social connection and engagement with robots.

While these studies tested the power of robotic attentiveness, we evaluate whether it can mediate the impact of competence on CT, thus exploring a more complex effect where multiple factors interact. By evaluating the combined impact of robot competence and attentiveness, we provide a broader view of the way it can shape interactions with robots.

\subsection{Robot competence}
Research on robotic competence has mainly focused on how robots' performance and reliability shape human perceptions and the outcomes of the interaction. For example, \citet{salem2015would} compared the effects of different error types on how participants perceived the robot. They contrasted cognitive mistakes (e.g., forgetting a person’s preferences) with physical errors (e.g., imprecise navigation) and reported that cognitive errors are especially damaging to the perception of robots \cite{salem2015would}. Robot competence was also indicated as an important factor that facilitates the robot's acceptance. \citet{natarajan2022effects} reported that capable robots were more easily accepted as team members in a game context. The robot's performance and human-likeness were the primary factors shaping its acceptance \cite{natarajan2022effects, short2010robot}. Recent work has also explored whether competence has a broad influence that carries over to affective aspects of the interaction. \citet{manor2025trust} investigated whether a robot's cognitive performance influences both CT and AT and reported that competence enhanced CT but did not affect AT. This asymmetric effect suggests that, although robots' capabilities build CT, competence alone is insufficient to form the emotional basis underlying AT.

The present work extends this line of research by investigating whether the effect of robotic competence depends on robot attentiveness. Whereas prior work tested competence in isolation, real‑world HRI often involves robots that vary in both cognitive capability and attentiveness. Thus, different combinations of capability and attentiveness may yield different outcomes for trust compared with those observed when each factor is studied alone.

%% file: section/3_methods.tex
\begin{table}[h]
\centering
\small 
\setlength{\tabcolsep}{5pt} 
\renewcommand{\arraystretch}{1.4} 
\begin{tabular}{|l|l|p{0.28\columnwidth}|p{0.28\columnwidth}|}
\hline
\multicolumn{2}{|c|}{} & \multicolumn{2}{c|}{\textbf{Attentiveness}} \\ \cline{3-4}
\multicolumn{2}{|c|}{} & \textbf{High} & \textbf{Low} \\ \hline

\multirow{3.5}{*}{\textbf{Competence}} & \textbf{High} &
  Finds cubes / Follows participant &
  Finds cubes / waits in place \\ \cline{2-4}
 & \textbf{Low} &
  Doesn't find cubes / Follows participant &
  Doesn't find cubes / Waits in place \\ \hline
\end{tabular}
\caption{Experimental conditions.}
\label{tab:attn_comp_wrap}
\end{table}

\section{Method}
We designed a human-robot collaborative task to investigate how the interplay between competence and attentiveness affects CT. Participants worked with a robotic dog, taking turns in a search task to achieve a shared goal. The robot's competence and attentiveness were manipulated in a $2\times2$ experimental design representing different combinations of the two variables (see Table \ref{tab:attn_comp_wrap}). Under each condition, the robot either found the relevant items during its search (high competence) or made frequent errors (low competence). It also either actively engaged with participants through attentive behavior (high attentiveness) or remained distant and unresponsive (low attentiveness). The study was approved by the ethics committee of the university.

\subsection{The Task}

\begin{figure}[b]
    \centering
\includegraphics[width=0.9\linewidth]{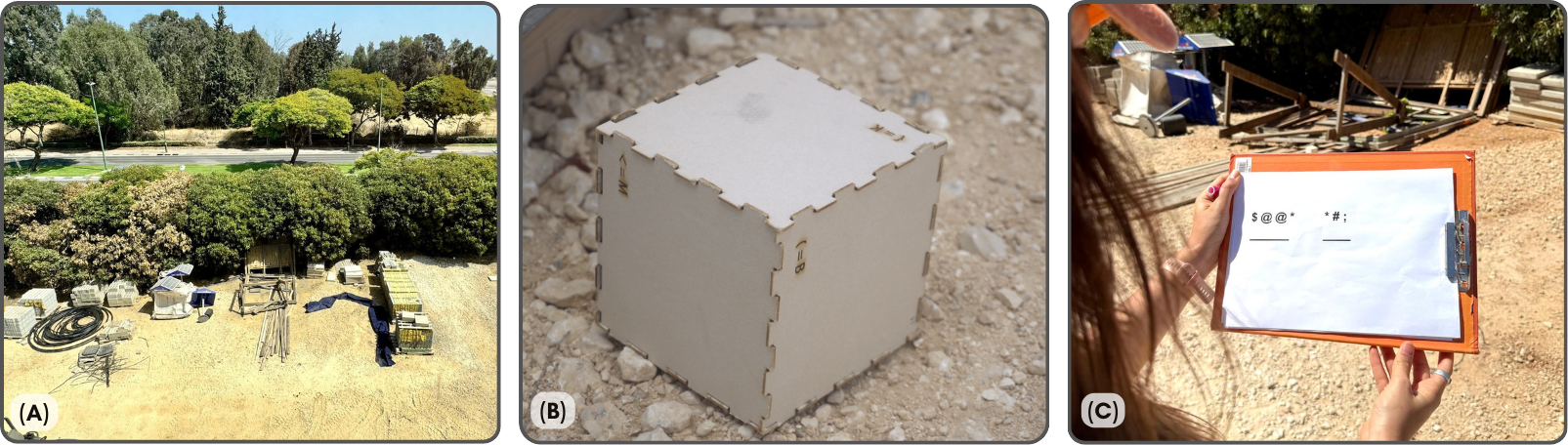}
   \caption{(A) The task's open area. (B) A cube with symbols. (C) The two-word code.}
 \label{TASK}
\end{figure}

The search task was based on a paradigm developed by \citet{carsenti2025raising} in which a participant teamed up with a robotic dog to search for cubes scattered in an open outdoor area (see Figure \ref{TASK}A). Written on the cubes was the translation of unfamiliar symbols used to compose a code (see Figure \ref{TASK}B). At the beginning of the task, participants were given a task sheet on which a sentence was written in the code. The sentence was composed of the symbols whose translations to letters appeared on the cubes. The task thus consisted of finding the cubes in the outdoor area and translating the symbols on the task sheet into letters (see Figure \ref{TASK}C). Participants were also informed that the cubes displayed various symbols, some of which were relevant to the code while others were not. We chose an open area with many physical features to create a suitable environment for the search task.

Ten cubes were spread across the $30\times30$ m$^2$ outdoor area. Each cube face (except the bottom) displayed an expression mapping an English letter to a symbol. For example, the notation *=D meant that the symbol * represented the letter D. To increase the task difficulty, only five of ten cubes had symbols relevant to the code, resulting in five different English letters. The two-word code \verb|$@@* *#;| translated into ``Good day'' when successfully deciphered (see Figure \ref{TASK}C).
The collaboration proceeded in turns, where on each turn, either the participant or the robot searched for a single cube and tested the symbols on it. When the robot located a cube, it would stand beside it and spin in place to attract the participant's attention, who would analyze the symbols on the cube to decipher the code. These alternating searches continued until the code was deciphered. The turn-based approach ensured collaboration and interdependence between the participant and the robot as both contributed to the task's completion.


\subsection{Design of robot behavior}
We used the Unitree Go2 robotic dog, a 15 kg quadruped with 12 degrees of freedom (see Figure \ref{fig:DAN}). We chose this robot because of its advanced capabilities, which differentiate it from the simpler robotic platforms used in previous HRI studies that investigated how competence or attentiveness affects CT. Unlike the simple lamp-like robot that was used in prior research \cite{manor2024attentiveness, hoffman2015design}, the robotic dog is a sophisticated autonomous system with multi-modal sensing and complex navigation capabilities, making it ideally suited for investigating HRI trust dynamics, which requires a competent robot that raises expectations for high competence and performance. Additionally, robotic dogs are naturally perceived as suitable for search and rescue tasks because they are associated with real dogs, which work collaboratively with humans in search scenarios \cite{carsenti2025raising}. The robotic dog is also allowed to avoid social complexities related to language and speech tone \cite{lakatos2014emotion, erel2024power, yaar2024performing}. The robot was operated remotely from within the building using the Wizard-of-Oz method, in which the researcher controlling the robot was hidden from sight \cite{mutlu2012conversational, riek2012wizard, carsenti2025raising}.

To manipulate the robot's competence and attentiveness, we designed specific robotic behaviors based on literature from psychology and HRI research. The robot's competence design was based on prior studies, which demonstrated that performance directly affects the perception of competence and reliability in collaborative contexts \cite{xu2018impact, salem2015would, manor2025trust}. \textbf{High competence behavior}, therefore, involved the robot's consistent identification of the relevant symbols on cubes, contributing meaningfully to deciphering the code. \textbf{Low competence behavior} included errors, such as the robot identifying cubes without the requisite symbols for deciphering the code or failing to identify cubes with relevant symbols, thus hindering task completion. The design of the robot's attentiveness was based on previous studies that identified relevant behavioral cues such as following, maintaining proximity, sharing emotional responses, and general responsiveness \cite{sidner2004study, moon2014meet, anzabi2023effect}. These nonverbal behaviors are known to signal interest, care, and willingness to assist, which together form a general sense of attentiveness \cite{manor2024attentiveness}. We therefore included these behavioral traits in the robot's \textbf{high attentive behavior} to demonstrate engagement, care, and responsiveness to the participant. For example, during the participant's turn to search, the robot followed the participant, maintaining close proximity. If the participant found a relevant cube, the robot responded with a celebratory ``happy dance'' characterized by energetic running and hopping in place to show enthusiasm for the participant's success. Conversely, if the participant found an irrelevant cube, the robot shared the participant's disappointment by shaking its head side to side (as if signaling ``no'') and then lowering its head. \textbf{Low attentive behavior} involved the opposite robotic behavior: the robot did not follow the participants, maintained a significant separation from them, and showed no responsiveness to the outcome of their search. The robot performed its own search independently, without demonstrating engagement or providing feedback regarding the participant's success or failure. 

Understanding and interpretation of these robotic behaviors were validated in a pilot study with 12 participants. All participants clearly understood the intended meaning of the gestures and reported the associated levels of competence and attentiveness.

Alongside the design of the robot’s behavior in the different conditions, we also designed a robotic greeting behavior to support a coherent interaction \cite{erel2024power}. At the beginning of each experiment (under all conditions), the robot introduced itself by approaching the participant, stopping approximately two meters away, and performing a greeting gesture by nodding its head up and down toward the participant. It then positioned itself beside the participant so that both faced the researcher, who provided instructions for the task. This behavior was designed to signal that the robot is partnering with the participant during the task and to comply with common social norms that establish positive interactions with robots \cite{carsenti2025raising, erel2024power}.


\subsection{Participants}
Eighty undergraduate students from the university's Communications, Psychology, and Computer Science departments participated in the study (57 females, 23 males; Mean age = 25.7, SD = 1.2). All participants completed an informed consent form and received course credits or shopping coupons for their participation.

\subsection{Experimental settings}
The experiment was conducted in an open area outside one of the university buildings. This area was quiet, surrounded by bushes and construction materials, providing suitable hiding places for the ten cubes (see Figure \ref{TASK}A). Participants began the experiment from a designated starting point, where they were given the task sheet with the code that they had to decipher. The robotic dog was initially hidden from view and approached the participants after they reached the starting point.

\subsection{Experimental design}

\begin{figure}[t]
    \centering
\includegraphics[width=0.7\linewidth]{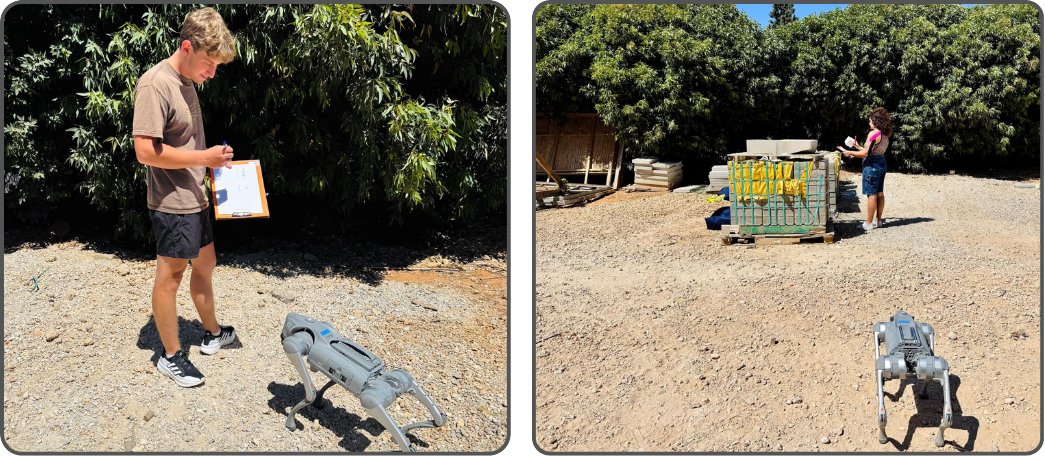}
   \caption{Robot attentive behaviors: (Left) high attentive behavior; (Right) low attentive behavior.}
 \label{CON}
\end{figure}

Using a between-participants $2\times2$ experimental design, we investigated the four possible combinations of competence (high vs. low) and attentiveness (high vs. low). The robot's competence and attentiveness were designed as follows:

\textbf{High competence--High attentiveness.} The robot finds cubes with relevant symbols. It follows the participants on their searches, shares their happiness (disappointment) when they find a cube with (without) relevant symbols (see Figure \ref{CON}, Left).

\textbf{High competence--Low attentiveness.} The robot finds cubes with relevant symbols. It does not follow the participants when they search, nor does it share their happiness (disappointment) when they find a cube with (without) relevant symbols (see Figure \ref{CON}, Right).

\textbf{Low competence--High attentiveness.} The robot either fails to find a cube or finds cubes with symbols that are irrelevant for deciphering the code. It only finds one cube that is relevant to the task. It follows the participants on their searches, shares their happiness (disappointment) when they find a cube with (without) relevant symbols (see Figure \ref{CON}, Left).

\textbf{Low competence--Low attentiveness}: The robot either fails to find a cube or finds cubes with symbols that are irrelevant for deciphering the code. It only finds one cube that is relevant to the task. It does not follow the participants when they search, nor does it share their happiness (disappointment) when they find a cube with (without) relevant symbols (see Figure \ref{CON}, Right).

Participants were randomly assigned to conditions using a matching technique that balanced gender and Negative Attitudes Toward Robots (NARS) \cite{nomura2006experimental} to avoid \textit{a priori} differences.

\subsection{Measures}
We used both objective and subjective measures to evaluate how the robot affected CT given the different combinations of robotic competence and attentiveness.

\subsubsection{Relying on the inaccurate recommendation of the robot in a cognitive task - performance on the Raven Progressive Matrices (RPMs)}

\begin{figure}[b]
    \centering
    \includegraphics[width=0.7\linewidth]{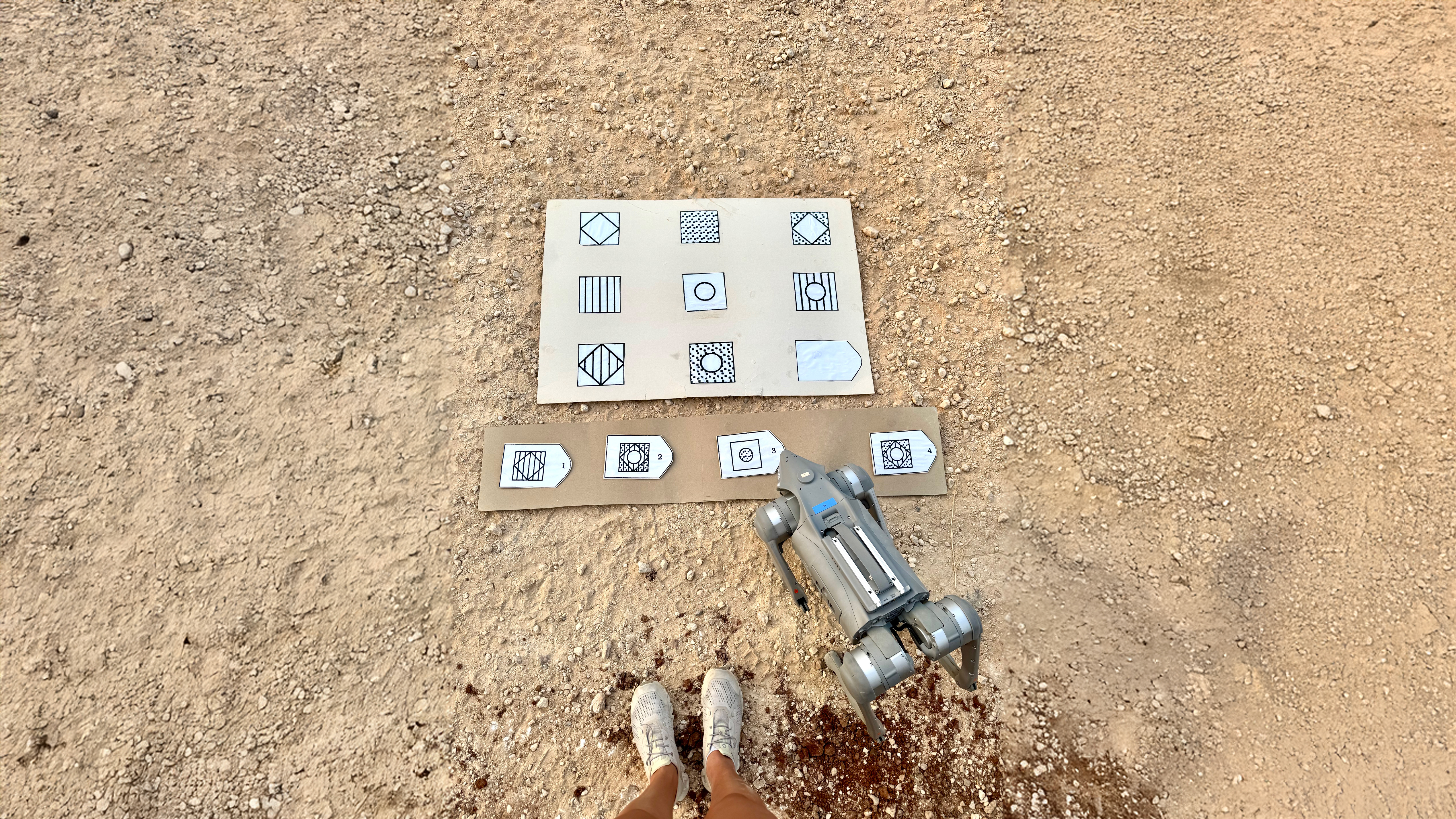}
    \caption{Physical version of RPM task.}
    \label{RAVEN}
\end{figure}

To assess whether the robot's competence and attentiveness during the search task affected CT, we investigated the participants' tendency to rely on the robot's recommendations in an unfamiliar cognitive task. We designed a physical task based on the RPMs, a widely used paradigm for assessing abstract reasoning capabilities \cite{raven1998raven}. Each item in the RPM comprises a $3\times3$ matrix of geometric shapes with one element missing. Participants were asked to identify the underlying rule governing the matrix and select the correct missing shape from a set of alternative answers. Although RPM tasks are conventionally administered via computer interfaces, we designed a large physical version tailored to the robot’s dimensions and communication capabilities (see Figure \ref{RAVEN}). This physical setup allowed the robot to engage in behaviors such as scanning the board and pointing to its recommended answer by standing next to it, thereby providing a visible and embodied indication of its recommendation for each item's answer.  

To mitigate potential fatigue effects, we used a reduced set of 9 items rather than the standard 18 items. The nine items were arranged in an ascending order of difficulty. Initial items involved straightforward rules, whereas subsequent items required the identification of multiple overlapping rules, spatial rotations, and abstract logical relationships. We also adjusted the number of alternative answers and modified the task to present four (instead of eight) alternatives for each item. This strategy made the robot’s selections easier to observe and reduced cognitive load.  

 Participants were informed that the robot was undergoing a training phase and was still learning to solve such tasks, thereby establishing the expectation that it might make errors. The participants were instructed to provide their answer for each item only after observing the robot’s recommendation. To investigate CT, we varied the robot's accuracy across the tasks, which progressively became increasingly difficult. The robot provided correct answers to the first three items, which were easy to solve. It then provided incorrect answers for items 4, 5, 6, 8, and 9. This distribution was designed to reflect competence on simpler items while simulating plausible errors on more challenging items, thereby creating a realistic learning scenario while avoiding the perception that the robot was malfunctioning (due to errors in elementary items). To evaluate CT, we measured participants' tendency to follow the robot's recommendation when it was inaccurate.

\subsubsection{Cognitive (and affective) trust scale}
We also assessed participants' self-reported CT by applying the adaptation of \citet{anzabi2023effect} for McAllister's trust scale \cite{mcallister1995affect}. The seven-point Likert scale (1 = strongly disagree, 7 = strongly agree) comprises nine statements that measure CT. Although not the primary focus of the study, we also included the seven statements that measure AT. 

\subsubsection{Perception of the robot's competence and warmth}
To further validate the competence and attentiveness manipulations, we used the competence and warmth sub-scales from the Robotic Social Attributes Scale (RoSAS), which is a 9-point Likert scale (1: ``low'' and 9: ``high'')  \cite{carpinella2017robotic}.

\subsubsection{Qualitative assessment}
A semi-structured interview was conducted to better understand participants' thoughts and attitudes \cite{boyatzis1998transforming}. The questions evaluated the participants' experience (e.g., ``describe the experience'') and their thoughts and perceptions about the robot (e.g., ``describe the robot'', or ``what did you think about it?''). 

\subsection{Procedure}
Participants received pretest questionnaires (NARS and demographics) by mail a few days prior to their arrival. Upon arrival at the designated outdoor search area, they received an overview of the experiment, specifying, among other things, that the experiment consisted of two phases and that in each phase they would be asked to perform a task (the search task and the RPM task).
Once a participant reached the starting point, the robot approached and performed a greeting gesture by walking toward the participant, nodding hello, and then positioning itself next to the participant. The researcher gave participants the task sheet and explained that their goal was to locate cubes that carry the translation for the symbols and decipher the code. The researcher also explained that the robot can recognize cubes and translate the symbols on them, but it is still in the training phase and may make mistakes. Participants were informed that the goal is to work together by taking turns, where, in each turn, either the participant or the robot examines a single cube. The researcher also explained that the robot will signal to the participant when it finds a relevant symbol during its turn. The researcher then left the participant with the robot to perform the task. The robotic behavior was activated according to the relevant conditions, and the task continued until the code was deciphered. After completing the search task, participants were invited to the second phase of the experiment. The researcher introduced the RPM task and explained that the robot was being trained to solve it and, therefore, might make mistakes. The participants were asked to wait for the robot to indicate its recommended answer for each item and only then give their own answer. Participants did not receive any feedback on their performance or that of the robot. 
Immediately after the participants completed the RPM task, they filled out the Cognitive (and Affective) Trust Questionnaire and the RoSAS Questionnaire. They then participated in an interview. At the end of the experiment, the researcher debriefed the participants and confirmed that they had an overall positive experience.

%% file: section/4_Analysis_and_Findings.tex
\section{Analysis}

To verify the lack of early differences between groups, we conducted a Bayesian analysis on the NARS pre-test. 

Our main analysis evaluated the impact of the two independent factors: (1) The robot's \textit{Competence} (high competence vs. low competence); and (2) The robot's \textit{Attentiveness} (high attentiveness vs. low attentiveness). Using a 2-way ANOVA, we tested the factors' mutual impact on the CT measures: (1) Percentage of participants' errors in the RPM that matched the robot's errors. Specifically, we coded participants' responses to the five items in which the robot provided an inaccurate recommendation and calculated the percentage of the items where participants chose the exact (inaccurate) answer recommended by the robot. This measure enabled us to evaluate CT because it indicated the extent to which participants followed the robot's recommendations even when it was inaccurate. (2) CT subscale; We conducted the same analysis on the AT subscale. We further applied a one-way ANOVA on the Competence and Warmth RoSAS subscales to validate the competence and attentiveness manipulations. The interviews were analyzed using thematic coding \cite{boyatzis1998transforming}, which involved three researchers who independently identified initial themes, discussed inconsistencies, and produced mutually agreed upon themes. They independently analyzed a subset of the interviews to verify inter-rater reliability ($\kappa=84\%$) and analyzed the remaining data.


 \section{Findings}


The Bayesian analysis revealed no early differences between conditions (NARS: $BF_{10} = 0.29$).
Our main analysis indicated a strong compensatory mechanism where high attentiveness compensated for low competence in the development of CT.

\subsection{Relying on the inaccurate recommendation of the robot in a cognitive task}
\begin{figure} [t]
    \centering
\includegraphics[width=0.72\linewidth]{Figures/RAVEN.png}
   \caption{Impact of robot competence and attentiveness on the percentage of items where the participant followed the robot's incorrect recommendations}
    \label{errors}
\end{figure}

\begin{figure} [b]
    \centering
\includegraphics[width=0.75\linewidth]{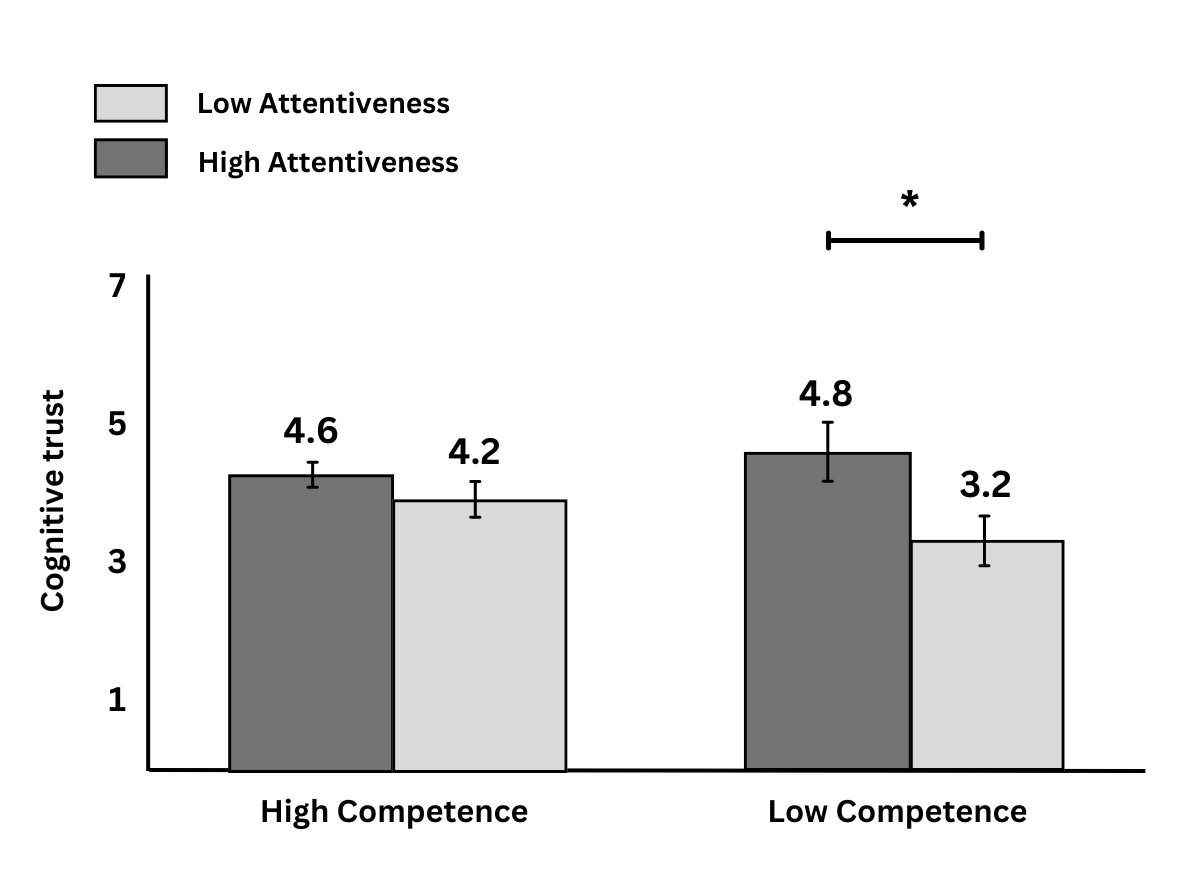}
  \caption{Impact of the robot competence and attentiveness on CT levels.}
    \label{GRAPH}
\end{figure}

The two-way ANOVA revealed a significant interaction between the robot's \textit{Competence} and its \textit{Attentiveness}, $F_{(1,76)} = 4.61,\; p = 0.035$. A Scheffe \textit{post hoc} analysis indicated that under \ high competence conditions, both high and low attentiveness resulted in a similar percentage of items where the participant followed the robot's inaccurate recommendation. Under low competence conditions, high attentiveness resulted in significantly higher percentages compared to low attentiveness ($p_{\text{contrast}} = 0.05$; see Figure \ref{errors}). Moreover, there were no differences between the percentages of following the robot's inaccurate recommendations in the high competence conditions and the low competence--high attentiveness condition. Hence, participants tended to rely on the robot's inaccurate recommendation (i.e., demonstrating high CT) after interacting with a competent robot (regardless of its attentiveness level) or after interacting with a highly attentive robot, even if its competence level was low. This finding indicated that robotic attentiveness compensated for low competence. The main effects were not significant.

\subsection{Cognitive trust subscale}
The two-way ANOVA analysis of the CT subscale revealed a similar interaction between the robot's \textit{Competence} and its \textit{Attentiveness}, $F_{(1,76)} = 12.96,\; p < 0.001$. Scheffe's \textit{post hoc} analysis indicated that, under the high competence conditions, both high and low attentiveness resulted in similar CT levels, whereas under the low competence conditions, high attentiveness resulted in significantly higher CT levels than low attentiveness ($p_{\text{contrast}}$ < 0.001; see Figure \ref{GRAPH}). This analysis also revealed no differences between the CT levels under the high competence conditions and the low competence--high attentiveness condition. These results further supported the possibility that high robotic attentiveness can compensate for low competence. The \textit{Competence} and \textit{Attentiveness} main effect were also significant, $\textit{F}_{(1,76)} = 6.76$, \textit{p} = 0.01; $\textit{F}_{(1,76)} = 34.04$, \textit{p} < 0.001.



\subsection{Attentive trust subscale}
The two-way ANOVA analysis of the AT subscale revealed a main effect of the robot's \textit{Attentiveness}, $F_{(1,76)} = 19.8,\; p < 0.001$, indicating that AT was greater under high attentiveness conditions  $M = 3.5,\; \text{SD} = 1.2$ than under low attentiveness conditions $M = 2.4,\; \text{SD} = 0.9$. All other effects were not significant.

\subsection{Validation of the manipulation}
Applying a one-way ANOVA to the RoSAS subscales validated our manipulations. The analysis of the competence subscale revealed a main effect for the robot's competence, $F_{(1,76)} = 21.65,\; p < 0.001$, indicating greater competence under high competence conditions ($M = 7.03,\; \text{SD} = 1.6$) than under low competence conditions (\textit{M = 5.5, SD = 2.6}). The analysis of the warmth subscale revealed an effect for the \textit{robot's attentiveness}, $\textit{F}_{(1,76)} = 74.86$, \textit{p} < 0.001, indicating higher warmth under high attentiveness conditions ($M = 6.8,\; \text{SD} = 1.3$) than under low attentiveness conditions ($M = 3.4,\; \text{SD} = 2.1$). 

\subsection {Qualitative analysis: Interview}
Altogether, 506 quotes were categorized into three main themes: cognitive trust, affective trust, and sense of collaboration. 

\subsubsection{Theme 1: Cognitive trust}
Cognitive trust was discussed by most participants (76/80). The level of trust varied between the different conditions. In the conditions that involved high competence, most participants (72\% high competence--high attentiveness, 68\% high competence--low attentiveness) reported high CT: \textit{``I can trust its judgment''} (P.55, F); \textit{``You can completely rely on its recommendations''} (P.63, F). The robots' flawless performance was a key reason for this trust: \textit{``It was always right''} (P.39, F); \textit{``It was never wrong''} (P.79, M). A minority of the participants (28\% high competence--high attentiveness, 32\% high competence--low attentiveness) expressed hesitation: \textit{``I had to double-check its performance''} (P.42, M). 

Participants in the low competence--high attentiveness condition also discussed CT, and a surprising majority (68\%) reported high levels of CT in the robot, despite its suboptimal performance. Unlike the high competence conditions, these participants explained their CT in social and emotional terms. Many described feeling a sense of safety: \textit{``I trusted it not because it's always right, but because it made me feel that I could count on it''} (P.32, M); others related to the robot's mistakes with empathy: \textit{``It was like me, I also do not get it right all the time''} (P. 63, F). Several participants were impressed by the robot's social skills, which exceeded their expectations: \textit{``It sensed me, it shows how sophisticated it is; you can definitely rely on it''} (P.62, F). However, 32\% expressed doubts about the robot’s reliability and reported low levels of trust: \textit{``It may not be as accurate as humans are. So you listen, but with caution''} (P.59, F). 

A different pattern emerged for the low competence--low attentiveness condition, where most participants (70\%) reported low levels of trust: \textit{``I could not rely on its suggestions''} (P.64, F); \textit{``I preferred to do it on my own''} (P.65, F). Their lack of trust was closely linked to the robot’s poor performance: \textit{``Once it got it wrong, I lost all confidence in it''} (P.34, M). Some also pointed out that they were better than the robot: \textit{``I found the cube long before it did''} (P.21, M). However, a smaller group (30\%) still reported high levels of trust: \textit{``It is probably right, it is very technological''} (P.54, M). 

\subsubsection{Theme 2: Affective trust}
Most participants (68/80) also discussed affective trust. Under conditions involving high attentiveness, the majority of participants reported high AT (81\% high competence--high attentiveness, 79\% low competence--high attentiveness). They reported a sense of connectedness: \textit{``This robot would be there for me''} (P.66, F); \textit{``It's there, with you''} (P.80, F), and explained that the robot made them feel comfortable: \textit{``It made me feel that it’s not embarrassing to make a mistake, it cared''} (P.67, F). A minority (19\% high competence--high attentiveness, 21\% low competence--high attentiveness) reported low AT: \textit{``I would not share emotions or personal things with it''} (P.43, M). 

Under the conditions that involved low attentiveness, most participants reported low AT (82\% high competence--low attentiveness, 81\% low competence--low attentiveness): \textit{``It was cold''} (P.78, F); \textit{``It's a machine, I wouldn't connect with it''} (P.15, M). Others wished the robot had been more attentive: \textit{``I wanted it to join me when I searched''} (P.5, M). A minority of the participants (18\% high competence-low attentiveness, 19\% low competence--low attentiveness conditions) reported higher levels of AT: \textit{``It was there helping me''} (P.38, M).   
 
\subsubsection{Theme 3: Sense of collaboration} 
Almost all participants (79/80) discussed the collaboration with the robot. In both high attentiveness conditions, most of the participants explicitly mentioned a high sense of collaboration (90\% high competence--high attentiveness, 75\% low competence--high attentiveness), attributing it to a sense of togetherness: \textit{``We were sharing a journey''} (P.44, F); \textit{``It felt like a partnership, we were trying together''} (P.32, M). The remaining participants (10\% high competence--high attentiveness, 25\% low competence--high attentiveness) described a lower sense of collaboration: \textit{``I could do it on my own, I didn’t need it''} (P.10, F). 

For the high competence--low attentiveness condition, 50\% of the participants reported that the robot's high level of competence strongly influenced their sense of collaboration: \textit{``It knew where and what to look for, we were working together''} (P.4, F). The other half of the participants described the robot as a dependent assistant: \textit{``The robot was dependent on me, I was in charge''} (P.5, M).

For the low competence--low attentiveness condition, only a few participants (15\%) mentioned a high sense of collaboration: \textit{``I felt it was a teammate''} (P.41, F). The vast majority (85\%) described a lack of cooperation, saying there was no real teamwork or connection: \textit{``There was no communication or collaboration''} (P.20, F); \textit{``The robot was doing its own thing, and so was I''} (P.64, F). 

%% file: section/5_discussion.tex
\section{Discussion}

In this study, we show that the dynamic between robots' competence and attentiveness plays a key role in shaping trust. Our findings reveal a powerful compensation mechanism in which robot attentiveness can establish CT, even when the robot performs poorly and directly hinders participants’ success. Specifically, high CT was observed in the two conditions involving high robotic competence, as predicted by traditional competence-based trust models. Interestingly, similar (high) CT levels were also observed when the robot demonstrated low competence and high attentiveness. Participants reported similarly high CT levels in the trust questionnaire and followed the robot's (incorrect) recommendations in a cognitive task to the same extent. The high level of CT under this condition suggests that the robot's competence did not automatically shape CT. Instead, its effect depended on the robot’s level of attentiveness. High attentiveness counteracted the negative impact of poor performance, leading to high CT, whereas low attentiveness resulted in a sharp decline.  

Two possible explanations for this compensatory effect emerged from participants' descriptions of their experience in the interviews. First, they explained that robots are typically perceived as task-oriented machines, so interacting with them is not expected to involve highly attentive behavior. The robot’s social capabilities thus exceeded expectations, and participants were impressed that the robot \textit{``came with''} them, \textit{``was there''} for them, and \textit{``cared about''} them. This appreciation of the robot's social capability led to the perception of the robot as \textit{``something you can count on''} and encouraged them to \textit{``trust its recommendations in cognitive tasks''}. The profound social impact of a robot that exceeds expectations has already been noted by previous studies \cite{erel2024rosi} and can explain the intense compensatory effect observed in the present study. Another explanation mentioned in the interviews related to a strong sense of empathy that emerged under conditions that involved high attentiveness. The robot's attentiveness triggered empathy and led to strong identification where participants described the robot as \textit{``just like me''} and \textit{``performing the task as I would.''} This identification encouraged participants to view the robot's mistakes as resembling human behavior, increasing their willingness to accept and understand them. They explained that the robot’s failure was reasonable, much like their own potential to miss a relevant cube: \textit{``It was trying its best. Just like me, it sometimes missed things or made mistakes. It still helped a lot''} (P.14, F). Rather than feeling disappointed, they related to the robot’s struggle, viewing its failure as legitimate and genuine \textit{``it showed me it’s okay not to be perfect---it makes mistakes, just like I do''} (P.62, F). These explanations suggest that the compensatory mechanism identified in our study involved an enhanced tolerance for errors that stemmed from identification with the robot and a feeling of interpersonal connection.

Taken together, these insights indicate that we should update traditional competence-based trust models, which predict a decline in CT following poor robot performance \cite{desai2012effects, hancock2011meta}. The present results suggest that CT formation in HRI may be more complex than previously understood and may involve a strong compensatory component triggered by emotional factors, whose effect on CT appears comparable to that of the robot’s competence. While high capabilities and reliability shape CT \cite{hancock2011meta, xu2018impact}, the impact of a robot's low competence depends on its attentiveness. Emotional bonds created through attentive behavior fundamentally alter how humans interpret and respond to poor robot performance and determine the level of trust during the interaction. Consider, for example, a human worker collaborating with a robotic arm in a big warehouse. Incorporating attentiveness into the robot's behavior (e.g., demonstration of lean, gaze, and cheerful gestures towards the worker at relevant times during their work together) may assist in shaping a positive perception and maintaining the worker's trust in the robotic arm, even if it makes occasional errors. This ability to preserve trust through attentiveness is particularly valuable for adaptive systems that learn through interactions with humans because maintaining trust during the early learning phase is crucial for long-term acceptance and adoption \cite{han2022systematic, hancock2011meta}. 

Another important perspective emerging from our findings concerns the central role of competence as a primary contributor to the development of CT. In line with previous research, we emphasize that a high level of competence is a strong foundation for trust. Once competence is established, it supports a stable sense of CT that is not easily undermined by other factors. The present results show that when the robot was reliable and made no errors, participants reported high CT even in the absence of attentiveness. Therefore, despite the importance of attentiveness as a compensatory mechanism that helps build CT, it is not essential for its formation. This pattern suggests that the decrease in trust reported in earlier studies under conditions of low attentiveness may be specific to situations where participants cannot adequately assess the robot’s competence.

More broadly, we highlight the need to explore more complex processes in the HRI domain. Interactions with robots are likely to involve multiple processes that occur simultaneously. Evaluating their combined effects rather than the influence of each in isolation may reveal valuable insights for designing robot behavior. We suggest that focusing on a single factor in HRI should be regarded as only a first step, to be followed by studies that identify additional factors that may mediate its influence. This approach will allow for a better representation of real-world HRIs and identifying mutual influences that should be carefully considered.





%% file: section/6_conclusion.tex
\section{Limitations}
This study has several limitations. First, we used a specific robot that performed a search task in a collaborative context. Future studies should investigate the dependency between competence and attentiveness using different robots performing different tasks in different contexts. Second, although participants showed a strong motivation to succeed in the search task and the RPM task, the results might have differed had these tasks carried meaningful consequences for them. Third, it is unclear whether competence and attentiveness would have similar effects when performing easier or harder tasks. Finally, qualitative findings may be influenced by the ``good-subject effect'' \cite{nichols2008good}, which we minimized by following a strict protocol and emphasizing that all responses were valuable.

\section{Conclusion}

This study extends our understanding of cognitive trust in HRI by adopting a comprehensive perspective that considers how competence and attentiveness jointly influence CT. Our findings revealed a powerful compensatory mechanism, where a high level of robot attentiveness can compensate for the negative effect of low robot competence in forming CT. This underexplored effect highlights the need to adjust the traditional emphasis on competence in models of CT and to account for the robot’s capacity for social and emotional engagement. We emphasize the importance of CT models that adopt this broader, human-centered perspective, suggesting that such models may be pivotal in sustaining effective and enduring HRI.